\documentclass[11pt,a4paper]{article}
\usepackage[hyperref]{acl2018}
\usepackage{times}
\usepackage{latexsym}

\usepackage{url}
\usepackage{amsmath,amssymb,amsthm}
\usepackage{color}
\usepackage{graphicx}
\usepackage{subcaption,siunitx,booktabs}
\usepackage{makecell}
\usepackage{multirow}
\usepackage{float}
\usepackage{arydshln}

\aclfinalcopy 
\def\aclpaperid{498} 

\newcommand{\newvec}[1]{\mathbf{#1}}
\DeclareMathOperator{\sigmoid}{sigmoid}

\title{Neural Coreference Resolution with Deep Biaffine Attention \\by Joint Mention Detection and Mention Clustering}

\author{Rui Zhang \thanks{\ \ Work done during the internship at IBM Watson.} \\ Yale University \\ r.zhang@yale.edu \And C\'{i}cero Nogueira dos Santos \\ IBM Research \\ cicerons@us.ibm.com \AND Michihiro  Yasunaga \\ Yale University \\ michihiro.yasunaga@yale.edu \And Bing Xiang \\ IBM Watson \\ bingxia@us.ibm.com \And Dragomir R. Radev \\ Yale University \\ dragomir.radev@yale.edu}

\date{}

\begin{document}
\maketitle
\begin{abstract}
Coreference resolution aims to identify in a text all mentions that refer to the same real-world entity.
The state-of-the-art end-to-end neural coreference model considers all text spans in a document as potential mentions and learns to link an antecedent for each possible mention.
In this paper, we propose to improve the end-to-end coreference resolution system by (1) using a biaffine attention model to get antecedent scores for each possible mention, and (2) jointly optimizing the mention detection accuracy and the mention clustering log-likelihood given the mention cluster labels.
Our model achieves the state-of-the-art performance on the CoNLL-2012 Shared Task English test set.
\end{abstract}

\section{Introduction}
End-to-end coreference resolution is the task of identifying and grouping \textit{mentions} in a text such that all mentions in a cluster refer to the same entity.
An example is given below \cite{bjorkelund2014learning} where mentions for two entities are labeled in two clusters:
\begin{quote}
[Drug Emporium Inc.]$_{a1}$ said [Gary Wilber]$_{b1}$ was named CEO of [this drugstore chain]$_{a2}$. [He]$_{b2}$ succeeds his father, Philip T. Wilber, who founded [the company]$_{a3}$ and remains chairman. Robert E. Lyons III, who headed the [company]$_{a4}$'s Philadelphia region, was appointed president and chief operating officer, succeeding [Gary Wilber]$_{b3}$.
\end{quote}
Many traditional coreference systems, either rule-based \cite{haghighi2009simple,lee2011stanford} or learning-based \cite{bengtson2008understanding,fernandes2012latent,durrett2013easy,bjorkelund2014learning}, usually solve the problem in two separate stages: (1) a mention detector to propose entity mentions from the text, and (2) a coreference resolver to cluster proposed mentions.
At both stages, they rely heavily on complicated, fine-grained, conjoined features via heuristics.
This pipeline approach can cause cascading errors, and in addition, since both stages rely on a syntactic parser and complicated hand-craft features, it is difficult to generalize to new data sets and languages.

Very recently, \newcite{lee2017end} proposed the first state-of-the-art end-to-end neural coreference resolution system.
They consider all text spans as potential mentions and therefore eliminate the need of carefully hand-engineered mention detection systems.
In addition, thanks to the representation power of pre-trained word embeddings and deep neural networks, the model only uses a minimal set of hand-engineered features (speaker ID, document genre, span distance, span width).
\begin{figure*}[ht!]
  \centering
  \includegraphics[clip,trim=0cm 1.5cm 0cm 1.4cm,width=\textwidth]{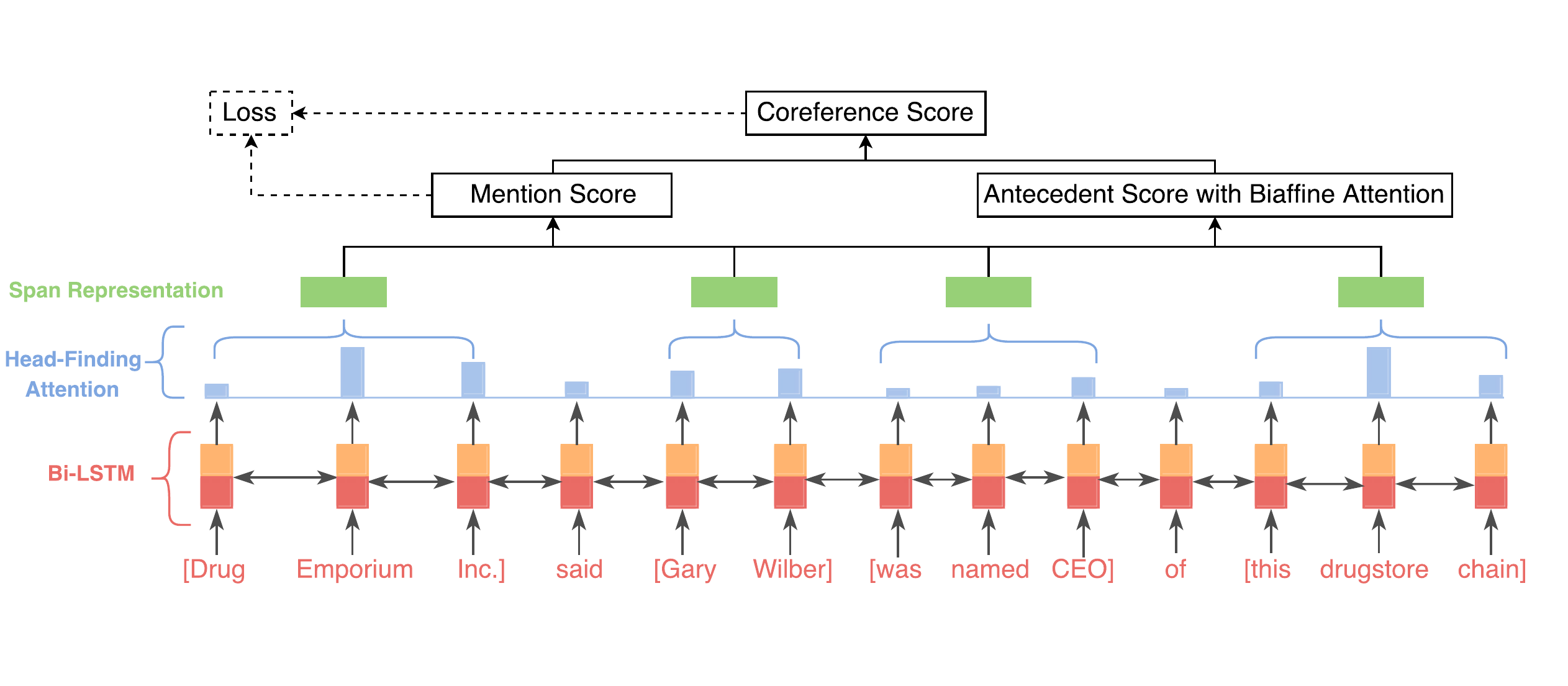}
  \caption{Model architecture. We consider all text spans up to 10-word length as possible mentions. For brevity, we only show three candidate antecedent spans (``Drug Emporium Inc.", ``Gary Wilber", ``was named CEO") for the current span ``this drugstore chain".}
  \label{fig:model}
\end{figure*}

The core of the end-to-end neural coreference resolver is the scoring function to compute the mention scores for all possible spans and the antecedent scores for a pair of spans.
Furthermore, one major challenge of coreference resolution is that most mentions in the document are singleton or non-anaphoric, i.e., not coreferent with any previous mention \cite{wiseman2015learning}.
Since the data set only have annotations for mention clusters, the end-to-end coreference resolution system needs to detect mentions, detect anaphoricity, and perform coreference linking.
Therefore, research questions still remain on good designs of the scoring architecture and the learning strategy for both mention detection and antecedent scoring given only the gold cluster labels.

To this end, we propose to use a biaffine attention model instead of pure feed forward networks to compute antecedent scores.
Furthermore, instead of training only to maximize the marginal likelihood of gold antecedent spans, we jointly optimize the mention detection accuracy and the mention clustering log-likelihood given the mention cluster labels.
We optimize mention detection loss explicitly to extract mentions and also perform anaphoricity detection.

We evaluate our model on the CoNLL-2012 English data set and achieve new state-of-the-art performances of 67.8\% F1 score using a single model and 69.2\% F1 score using a 5-model ensemble.

\section{Task Formulation}
In end-to-end coreference resolution, the input is a document $D$ with $T$ words, and the output is a set of mention clusters each of which refers to the same entity.
A possible \textit{span} is an N-gram within a single sentence.
We consider all possible spans up to a predefined maximum width. 
To impose an ordering, spans are sorted by the start position $\text{START}(i)$ and then by the end position $\text{END}(i)$.
For each span $i$ the system needs to assign an antecedent $a_i$ from all preceding spans or a dummy antecedent $\epsilon$:
$a_i \in \{\epsilon,1,\dots,i-1\}$.
If a span $j$ is a true antecedent of the span $i$, then we have $a_i=j$ and $1 \leq j \leq i-1$.
The dummy antecedent $\epsilon$ represents two possibilities: (1) the span $i$ is not an entity mention, or (2) the span $i$ is an entity mention but not coreferent with any previous span.
Finally, the system groups mentions according to coreference links to form the mention clusters.

\section{Model}
Figure \ref{fig:model} illustrates our model.
We adopt the same span representation approach as in \newcite{lee2017end} using bidirectional LSTMs and a head-finding attention.
Thereafter, a feed forward network produces scores for spans being entity mentions.
For antecedent scoring, we propose a biaffine attention model \cite{dozat2017deep} to produce distributions of possible antecedents.
Our training data only provides gold mention cluster labels.
To make best use of this information, we propose to jointly optimize the mention scoring and antecedent scoring in our loss function.\\
\textbf{Span Representation}
Suppose the current sentence of length $L$ is $[w_1,w_2,\dots,w_L]$, we use $\newvec{w}_t$ to denote the concatenation of fixed pretrained word embeddings and CNN character embeddings \cite{santos2014:icml} for word $w_t$.
Bidirectional LSTMs \cite{hochreiter1997long} recurrently encode each $w_t$:
\begin{align}
\label{eq:bilstm}
\begin{split}
   \overrightarrow{\newvec{h}}_{t} &= \mathrm{LSTM}^{\text{forward}}(\overrightarrow{\newvec{h}}_{t-1},\newvec{w}_{t})\\
   \overleftarrow{\newvec{h}}_{t} &= \mathrm{LSTM}^{\text{backward}}(\overleftarrow{\newvec{h}}_{t+1},\newvec{w}_{t})\\
   \newvec{h}_{t} & = [\overrightarrow{\newvec{h}}_{t},\overleftarrow{\newvec{h}}_{t}]
\end{split}
\end{align}
Then, the head-finding attention computes a score distribution over different words in a span $s_i$:
\begin{align}
\label{eq:headatt}
\begin{split}
\alpha_t &= \newvec{v}_{\alpha}^{\intercal}\mathrm{FFNN}_{\alpha}(\newvec{h}_{t}) \\
s_{i,t} &= \frac{\exp(\alpha_t)}{\sum\limits_{k=\text{START}(i)}^{\text{END}(i)}\exp(\alpha_k)} \\
\newvec{w}_{i}^{\text{head-att}} &= \sum_{t=\text{START}(i)}^{\text{END}(i)}s_{i,t}\newvec{w}_{t}\\
\end{split}
\end{align}
where $\mathrm{FFNN}$ is a feed forward network outputting a vector.
\begin{table*}[t!]
\centering
\small
\begin{tabular}{ccccccccccc}
                  & \multicolumn{3}{c}{MUC} & \multicolumn{3}{c}{B$^{3}$} & \multicolumn{3}{c}{CEAF$_{\phi_4}$} &     \\
                  & P & R & F1 & P & R & F1 & P & R & F1 & Avg. F1 \\ \hline
\textbf{Our work (5-model ensemble)} & 82.1 & 73.6 & 77.6 & 73.1 & 62.0 & 67.1 & 67.5 & 59.0 & 62.9 & \textbf{69.2} \\
\newcite{lee2017end} (5-model ensemble)  & 81.2 & 73.6 & 77.2 & 72.3 & 61.7 & 66.6 & 65.2 & 60.2 & 62.6 & 68.8 \\ \hdashline
\textbf{Our work (single model)}    & 79.4 & 73.8 & 76.5 & 69.0 & 62.3 & 65.5 & 64.9 & 58.3 & 61.4 & \textbf{67.8} \\
\newcite{lee2017end} (single model)    & 78.4 & 73.4 & 75.8 & 68.6 & 61.8 & 65.0 & 62.7 & 59.0 & 60.8 & 67.2 \\
\newcite{clark2016deep}          & 79.2 & 70.4 & 74.6 & 69.9 & 58.0 & 63.4 & 63.5 & 55.5 & 59.2 & 65.7 \\
\newcite{clark2016improving}     & 79.9 & 69.3 & 74.2 & 71.0 & 56.5 & 63.0 & 63.8 & 54.3 & 58.7 & 65.3 \\
\newcite{wiseman2016learning}    & 77.5 & 69.8 & 73.4 & 66.8 & 57.0 & 61.5 & 62.1 & 53.9 & 57.7 & 64.2 \\
\newcite{wiseman2015learning}    & 76.2 & 69.3 & 72.6 & 66.2 & 55.8 & 60.5 & 59.4 & 54.9 & 57.1 & 63.4 \\
\newcite{fernandes2014latent} 	 & 75.9 & 65.8 & 70.5 & 77.7 & 65.8 & 71.2 & 43.2 & 55.0 & 48.4 & 63.4 \\
\newcite{clark2015entity}        & 76.1 & 69.4 & 72.6 & 65.6 & 56.0 & 60.4 & 59.4 & 53.0 & 56.0 & 63.0 \\
\newcite{martschat2015latent}    & 76.7 & 68.1 & 72.2 & 66.1 & 54.2 & 59.6 & 59.5 & 52.3 & 55.7 & 62.5 \\
\newcite{durrett2014joint}       & 72.6 & 69.9 & 71.2 & 61.2 & 56.4 & 58.7 & 56.2 & 54.2 & 55.2 & 61.7 \\
\newcite{bjorkelund2014learning} & 74.3 & 67.5 & 70.7 & 62.7 & 55.0 & 58.6 & 59.4 & 52.3 & 55.6 & 61.6 \\
\newcite{durrett2013easy}        & 72.9 & 65.9 & 69.2 & 63.6 & 52.5 & 57.5 & 54.3 & 54.4 & 54.3 & 60.3 \\
\end{tabular}
\caption{Experimental results on the CoNLL-2012 Englisth test set. The F1 improvements are statistical significant with $p < 0.05$ under the paired bootstrap resample test \cite{koehn2004statistical} compared with \newcite{lee2017end}.}
\label{tab:result_conll2012}
\end{table*}

Effective span representations encode both contextual information and internal structure of spans.
Therefore, we concatenate different vectors, including a feature vector $\phi(i)$ for the span size, to produce the span representation $\newvec{s}_i$ for $s_i$:
\begin{equation}
    \newvec{s}_i = [\newvec{h}_{\text{START}(i)},\newvec{h}_{\text{END}(i)},\newvec{w}_{i}^{\text{head-att}},\phi(i)]
\end{equation}
\textbf{Mention Scoring} The span representation is input to a feed forward network which measures if it is an entity mention using a score $m(i)$:
\begin{equation}
  m(i) = \newvec{v}_{\text{m}}^{\intercal}\mathrm{FFNN}_{\text{m}}(\newvec{s}_i)
\end{equation}
Since we consider all possible spans, the number of spans is $O(T^2)$ and the number of span pairs is $O(T^4)$.
Due to computation efficiency, we prune candidate spans during both inference and training.
We keep $\lambda T$ spans with highest mention scores.\\
\textbf{Biaffine Attention Antecedent Scoring}
Consider the current span $s_i$ and its previous spans $s_j$ ($1 \leq j \leq i-1$), we propose to use a biaffine attention model to produce scores $c(i,j)$:
\begin{align}
\label{eq:biaffine}
\begin{split}
  \hat{\newvec{s}}_{\text{i}} &= \mathrm{FFNN}_{\text{anaphora}}(\newvec{s}_{\text{i}})\\
  \hat{\newvec{s}}_{\text{j}} &= \mathrm{FFNN}_{\text{antecedent}}(\newvec{s}_{\text{j}}), 1 \leq j \leq i-1\\
  c(i,j) &= \hat{\newvec{s}}_{\text{j}}^{\intercal}\newvec{U}_{\text{bi}}\hat{\newvec{s}}_{\text{i}} + \newvec{v}_{\text{bi}}^{\intercal}\hat{\newvec{s}}_{\text{i}}\\
\end{split}
\end{align}
$\mathrm{FFNN}_{\text{anaphora}}$ and $\mathrm{FFNN}_{\text{antecedent}}$ reduce span representation dimensions and only keep information relevant to coreference decisions.
Compared with the traditional $\mathrm{FFNN}$ approach in \newcite{lee2017end}, biaffine attention directly models both the compatibility of $s_i$ and $s_j$ by $\hat{\newvec{s}}_{\text{j}}^{\intercal}\newvec{U}_{\text{bi}}\hat{\newvec{s}}_{\text{i}}$ and the prior likelihood of $s_i$ having an antecedent by $\newvec{v}_{\text{bi}}^{\intercal}\hat{\newvec{s}}_{\text{i}}$.\\
\textbf{Inference}
The final coreference score $s(i,j)$ for span $s_i$ and span $s_j$ consists of three terms: (1) if $s_i$ is a mention, (2) if $s_j$ is a mention, (3) if $s_j$ is an antecedent for $s_i$.
Furthermore, for dummy antecedent $\epsilon$, we fix the final score to be 0:
\begin{equation}
    s(i,j) =
\begin{cases}
    m(i)+m(j)+c(i,j), & j \neq \epsilon \\
    0, &j = \epsilon \\
\end{cases}
\end{equation}
During inference, the model only creates a link if the highest antecedent score is positive.\\
\textbf{Joint Mention Detection and Mention Cluster}
During training, only mention cluster labels are available rather than antecedent links.
Therefore, \newcite{lee2017end} train the model end-to-end by maximizing the following marginal log-likelihood where $\text{GOLD}(i)$ are gold antecedents for $s_i$:
\begin{align}
\begin{split}
    \mathcal{L}_{\text{cluster}}(i) &= \log \frac{\sum_{j^{\prime}\in \text{GOLD}(i)}\exp(s(i,j^{\prime}))}{\sum_{j=\epsilon,0,\dots,i-1} \exp(s(i,j))} \\
\end{split}
\end{align}
However, the initial pruning is completely random and the mention scoring model only receives distant supervision if we only optimize the above mention cluster performance.
This makes learning slow and ineffective especially for mention detection.
Based on this observation, we propose to directly optimize mention detection:
\begin{align}
\begin{split}
    \mathcal{L}_{\text{detect}}(i) &=  y_i\log\hat{y}_i + (1-y_i)\log(1-\hat{y}_i)
\end{split}
\end{align}
where $\hat{y}_i=\sigmoid(m(i))$, $y_i=1$ if and only if $s_i$ is in one of the gold mention clusters.
Our final loss combines mention detection and clustering:
\begin{equation*}
    \mathcal{L}_{\text{loss}} = -\lambda_{\text{detect}}\sum_{i=1}^{N}\mathcal{L}_{\text{detect}}(i) - \sum_{i^{\prime}=1}^{N^{\prime}}\mathcal{L}_{\text{cluster}}(i^{\prime})
\end{equation*}
where $N$ is the number of all possible spans, $N^{\prime}$ is the number of unpruned spans, and $\lambda_{\text{detection}}$ controls weights of two terms.

\section{Experiments}
\textbf{Data Set and Evaluation}
We evaluate our model on the CoNLL-2012 Shared Task English data \cite{pradhan2012conll} which is based on the OntoNotes corpus \cite{hovy2006ontonotes}.
It contains 2,802/343/348 train/development/test documents in different genres.

We use three standard metrics: MUC \cite{vilain1995model}, B$^{3}$ \cite{bagga1998algorithms}, and CEAF$_{\phi_4}$ \cite{luo2005coreference}.
We report Precision, Recall, F1 for each metric and the average F1 as the final CoNLL score.\\
\textbf{Implementation Details} For fair comparisons, we follow the same hyperparameters as in \newcite{lee2017end}.
We consider all spans up to 10 words and up to 250 antecedents.
$\lambda=0.4$ is used for span pruning.
We use fixed concatenations of 300-dimension GloVe \cite{pennington2014glove} embeddings and 50-dimension embeddings from \newcite{turian2010word}.
Character CNNs use 8-dimension learned embeddings and 50 kernels for each window size in \{3,4,5\}.
LSTMs have hidden size 200, and each $\mathrm{FFNN}$ has two hidden layers with 150 units and ReLU \cite{nair2010rectified} activations.
We include (speaker ID, document genre, span distance, span width) features as 20-dimensional learned embeddings.
Word and character embeddings use 0.5 dropout.
All hidden layers and feature embeddings use 0.2 dropout.
The batch size is 1 document.
Based on the results on the development set, $\lambda_{\text{detection}}=0.1$ works best from $\{0.05,0.1,0.5,1.0\}$.
Model is trained with ADAM optimizer \cite{kingma2014adam} and converges in around 200K updates, which is faster than that of \newcite{lee2017end}.
\begin{table}[t]
\centering
\begin{tabular}{lc}
                               & Avg. F1 \\ \hline
Our model (single)             & 67.8    \\
without mention detection loss & 67.5    \\
without biaffine attention     & 67.4    \\ \hdashline
\newcite{lee2017end}           & 67.3    \\
\end{tabular}
\caption{Ablation study on the development set.}
\label{tab:ablation}
\end{table}
\begin{figure}[t]
  \centering
  \includegraphics[clip,trim=0.2cm 0.3cm 0cm 0.1cm,width=.48\textwidth]{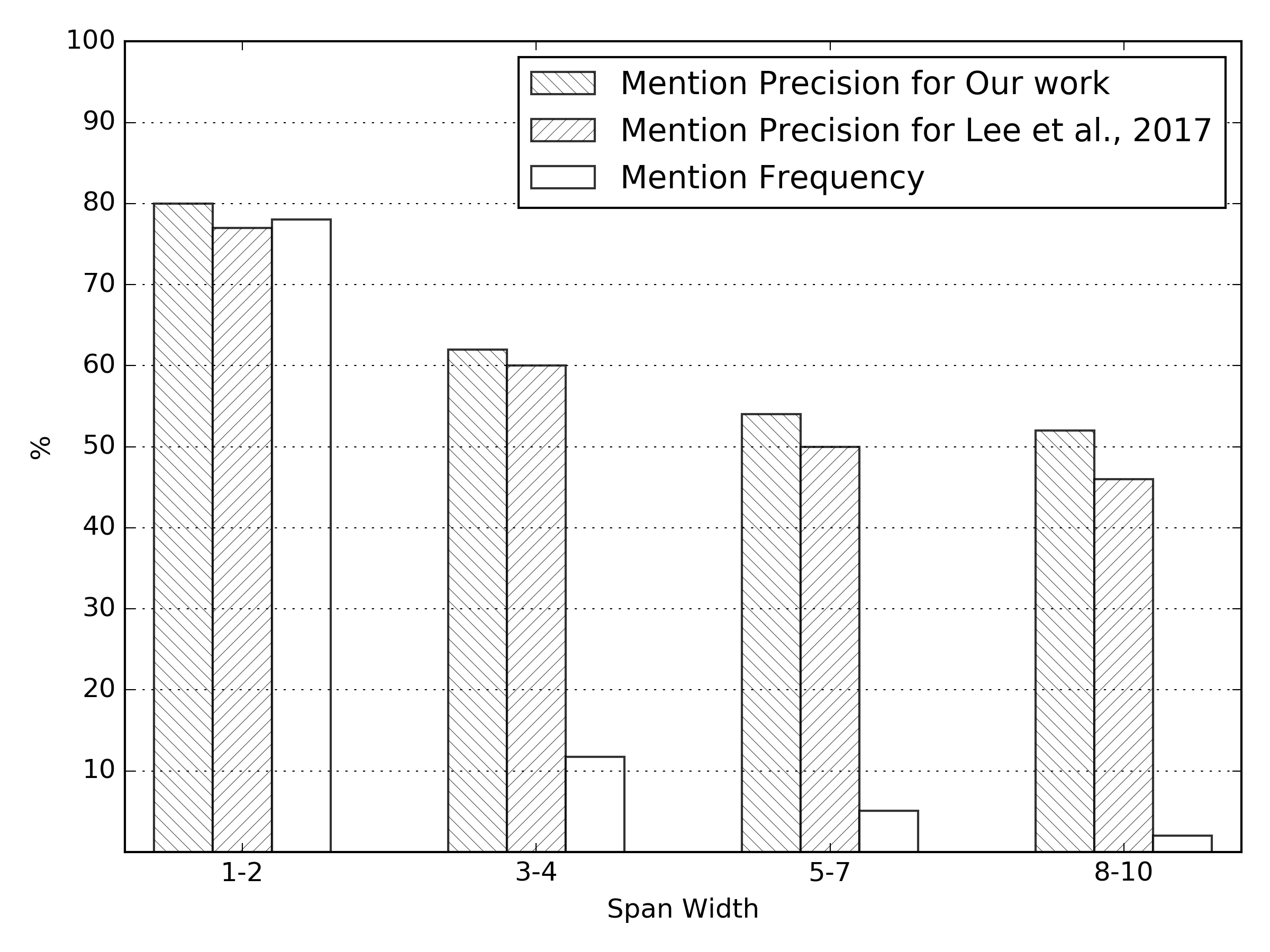}
  \caption{Mention detection subtask on development set. We plot accuracy and frequency breakdown by span widths.}
  \label{fig:mention_acc}
\end{figure}
\\\textbf{Overall Performance}
In Table \ref{tab:result_conll2012}, we compare our model with previous state-of-the-art systems.
We obtain the best results in all F1 metrics.
Our single model achieves 67.8\% F1 and our 5-model ensemble achieves 69.2\% F1.
In particular, compared with \newcite{lee2017end}, our improvement mainly results from the precision scores.
This indicates that the mention detection loss does produce better mention scores and the biaffine attention more effectively determines if two spans are coreferent.\\
\textbf{Ablation Study} To understand the effect of different proposed components, we perform ablation study on the development set.
As shown in Table \ref{tab:ablation}, removing the mention detection loss term or the biaffine attention decreases 0.3/0.4 final F1 score, but still higher than the baseline.
This shows that both components have contributions and when they work together the total gain is even higher.\\
\textbf{Mention Detection Subtask} To further understand our model, we perform a mention detection subtask where spans with mention scores higher than 0 are considered as mentions.
We show the mention detection accuracy breakdown by span widths in Figure \ref{fig:mention_acc}.
Our model indeed performs better thanks to the mention detection loss.
The advantage is even clearer for longer spans which consist of 5 or more words.

In addition, it is important to note that our model can detect mentions that do not exist in the training data. 
While \newcite{moosavi2017lexical} observe that there is a large overlap between the gold mentions of the training and dev (test) sets, we find that our model can correctly detect 1048 mentions which are not detected by \newcite{lee2017end}, consisting of 386 mentions existing in training data and 662 mentions not existing in training data.
From those 662 mentions, some examples are (1) a suicide murder (2) Hong Kong Island (3) a US Airforce jet carrying robotic undersea vehicles (4) the investigation into who was behind the apparent suicide attack.
This shows that our mention loss helps detection by generalizing to new mentions in test data rather than memorizing the existing mentions in training data.

\section{Related Work}
As summarized by \newcite{ng2010supervised}, learning-based coreference models can be categorized into three types:
(1) Mention-pair models train binary classifiers to determine if a pair of mentions are coreferent \cite{soon2001machine,ng2002improving,bengtson2008understanding}.
(2) Mention-ranking models explicitly rank all previous candidate mentions for the current mention and select a single highest scoring antecedent for each anaphoric mention \cite{denis2007ranking,wiseman2015learning,clark2016deep,lee2017end}.
(3) Entity-mention models learn classifiers to determine whether the current mention is coreferent with a preceding, partially-formed mention cluster \cite{clark2015entity,wiseman2016learning,clark2016improving}.

In addition, we also note latent-antecedent models \cite{fernandes2012latent,bjorkelund2014learning,martschat2015latent}.
\newcite{fernandes2012latent} introduce coreference trees to represent mention clusters and learn to extract the maximum scoring tree in the graph of mentions.

Recently, several neural coreference resolution systems have achieved impressive gains \cite{wiseman2015learning,wiseman2016learning,clark2016improving,clark2016deep}.
They utilize distributed representations of mention pairs or mention clusters to dramatically reduce the number of hand-crafted features.
For example, \newcite{wiseman2015learning} propose the first neural coreference resolution system by training a deep feed-forward neural network for mention ranking.
However, these models still employ the two-stage pipeline and require a syntactic parser or a separate designed hand-engineered mention detector.

Finally, we also note the relevant work on joint mention detection and coreference resolution.
\newcite{daume2005large} propose to model both mention detection and coreference of the Entity Detection and Tracking task simultaneously.
\newcite{denis2007joint} propose to use integer linear programming framework to model anaphoricity and coreference as a joint task.

\section{Conclusion}
In this paper, we propose to use a biaffine attention model to jointly optimize mention detection and mention clustering in the end-to-end neural coreference resolver.
Our model achieves the state-of-the-art performance on the CoNLL-2012 Shared Task in English.
\section*{Acknowledgments}
We thank Kenton Lee and three anonymous reviewers for their helpful discussion and feedback.

\bibliography{acl2018}

\begin{thebibliography}{35}
\expandafter\ifx\csname natexlab\endcsname\relax\def\natexlab#1{#1}\fi

\bibitem[{Bagga and Baldwin(1998)}]{bagga1998algorithms}
Amit Bagga and Breck Baldwin. 1998.
\newblock Algorithms for scoring coreference chains.
\newblock In \emph{The first international conference on language resources and
  evaluation workshop on linguistics coreference}.

\bibitem[{Bengtson and Roth(2008)}]{bengtson2008understanding}
Eric Bengtson and Dan Roth. 2008.
\newblock Understanding the value of features for coreference resolution.
\newblock In \emph{EMNLP}.

\bibitem[{Bj{\"o}rkelund and Kuhn(2014)}]{bjorkelund2014learning}
Anders Bj{\"o}rkelund and Jonas Kuhn. 2014.
\newblock Learning structured perceptrons for coreference resolution with
  latent antecedents and non-local features.
\newblock In \emph{ACL}.

\bibitem[{Clark and Manning(2015)}]{clark2015entity}
Kevin Clark and Christopher~D. Manning. 2015.
\newblock Entity-centric coreference resolution with model stacking.
\newblock In \emph{ACL}.

\bibitem[{Clark and Manning(2016{\natexlab{a}})}]{clark2016deep}
Kevin Clark and Christopher~D. Manning. 2016{\natexlab{a}}.
\newblock Deep reinforcement learning for mention-ranking coreference models.
\newblock In \emph{EMNLP}.

\bibitem[{Clark and Manning(2016{\natexlab{b}})}]{clark2016improving}
Kevin Clark and Christopher~D. Manning. 2016{\natexlab{b}}.
\newblock Improving coreference resolution by learning entity-level distributed
  representations.
\newblock In \emph{ACL}.

\bibitem[{Daum\'e~III and Marcu(2005)}]{daume2005large}
Hal Daum\'e~III and Daniel Marcu. 2005.
\newblock A large-scale exploration of effective global features for a joint
  entity detection and tracking model.
\newblock In \emph{Proceedings of Human Language Technology Conference and
  Conference on Empirical Methods in Natural Language Processing}.

\bibitem[{Denis and Baldridge(2007{\natexlab{a}})}]{denis2007joint}
Pascal Denis and Jason Baldridge. 2007{\natexlab{a}}.
\newblock Joint determination of anaphoricity and coreference resolution using
  integer programming.
\newblock In \emph{NAACL}.

\bibitem[{Denis and Baldridge(2007{\natexlab{b}})}]{denis2007ranking}
Pascal Denis and Jason Baldridge. 2007{\natexlab{b}}.
\newblock A ranking approach to pronoun resolution.
\newblock In \emph{IJCAI}.

\bibitem[{Dozat and Manning(2017)}]{dozat2017deep}
Timothy Dozat and Christopher~D Manning. 2017.
\newblock Deep biaffine attention for neural dependency parsing.
\newblock In \emph{ICLR}.

\bibitem[{Durrett and Klein(2013)}]{durrett2013easy}
Greg Durrett and Dan Klein. 2013.
\newblock Easy victories and uphill battles in coreference resolution.
\newblock In \emph{EMNLP}.

\bibitem[{Durrett and Klein(2014)}]{durrett2014joint}
Greg Durrett and Dan Klein. 2014.
\newblock A joint model for entity analysis: Coreference, typing, and linking.
\newblock \emph{Transactions of the Association for Computational Linguistics},
  2:477--490.

\bibitem[{Fernandes et~al.(2012)Fernandes, Dos~Santos, and
  Milidi{\'u}}]{fernandes2012latent}
Eraldo~Rezende Fernandes, C{\'\i}cero~Nogueira Dos~Santos, and Ruy~Luiz
  Milidi{\'u}. 2012.
\newblock Latent structure perceptron with feature induction for unrestricted
  coreference resolution.
\newblock In \emph{Joint Conference on EMNLP and CoNLL-Shared Task}.

\bibitem[{Fernandes et~al.(2014)Fernandes, dos Santos, and
  Milidi{\'u}}]{fernandes2014latent}
Eraldo~Rezende Fernandes, C{\'\i}cero~Nogueira dos Santos, and Ruy~Luiz
  Milidi{\'u}. 2014.
\newblock Latent trees for coreference resolution.
\newblock \emph{Computational Linguistics}.

\bibitem[{Haghighi and Klein(2009)}]{haghighi2009simple}
Aria Haghighi and Dan Klein. 2009.
\newblock Simple coreference resolution with rich syntactic and semantic
  features.
\newblock In \emph{EMNLP}.

\bibitem[{Hochreiter and Schmidhuber(1997)}]{hochreiter1997long}
Sepp Hochreiter and J{\"u}rgen Schmidhuber. 1997.
\newblock Long short-term memory.
\newblock \emph{Neural computation}, 9(8):1735--1780.

\bibitem[{Hovy et~al.(2006)Hovy, Marcus, Palmer, Ramshaw, and
  Weischedel}]{hovy2006ontonotes}
Eduard Hovy, Mitchell Marcus, Martha Palmer, Lance Ramshaw, and Ralph
  Weischedel. 2006.
\newblock Ontonotes: the 90\% solution.
\newblock In \emph{NAACL}.

\bibitem[{Kingma and Ba(2015)}]{kingma2014adam}
Diederik Kingma and Jimmy Ba. 2015.
\newblock Adam: A method for stochastic optimization.
\newblock In \emph{ICLR}.

\bibitem[{Koehn(2004)}]{koehn2004statistical}
Philipp Koehn. 2004.
\newblock Statistical significance tests for machine translation evaluation.
\newblock In \emph{EMNLP}.

\bibitem[{Lee et~al.(2011)Lee, Peirsman, Chang, Chambers, Surdeanu, and
  Jurafsky}]{lee2011stanford}
Heeyoung Lee, Yves Peirsman, Angel Chang, Nathanael Chambers, Mihai Surdeanu,
  and Dan Jurafsky. 2011.
\newblock Stanford's multi-pass sieve coreference resolution system at the
  conll-2011 shared task.
\newblock In \emph{Proceedings of the fifteenth conference on computational
  natural language learning: Shared task}, pages 28--34.

\bibitem[{Lee et~al.(2017)Lee, He, Lewis, and Zettlemoyer}]{lee2017end}
Kenton Lee, Luheng He, Mike Lewis, and Luke Zettlemoyer. 2017.
\newblock End-to-end neural coreference resolution.
\newblock In \emph{EMNLP}.

\bibitem[{Luo(2005)}]{luo2005coreference}
Xiaoqiang Luo. 2005.
\newblock On coreference resolution performance metrics.
\newblock In \emph{EMNLP}.

\bibitem[{Martschat and Strube(2015)}]{martschat2015latent}
Sebastian Martschat and Michael Strube. 2015.
\newblock Latent structures for coreference resolution.
\newblock \emph{Transactions of the Association for Computational Linguistics},
  3:405--418.

\bibitem[{Moosavi and Strube(2017)}]{moosavi2017lexical}
Nafise~Sadat Moosavi and Michael Strube. 2017.
\newblock Lexical features in coreference resolution: To be used with caution.
\newblock In \emph{ACL}.

\bibitem[{Nair and Hinton(2010)}]{nair2010rectified}
Vinod Nair and Geoffrey~E Hinton. 2010.
\newblock Rectified linear units improve restricted boltzmann machines.
\newblock In \emph{ICML}.

\bibitem[{Ng(2010)}]{ng2010supervised}
Vincent Ng. 2010.
\newblock Supervised noun phrase coreference research: The first fifteen years.
\newblock In \emph{ACL}.

\bibitem[{Ng and Cardie(2002)}]{ng2002improving}
Vincent Ng and Claire Cardie. 2002.
\newblock Improving machine learning approaches to coreference resolution.
\newblock In \emph{ACL}.

\bibitem[{Pennington et~al.(2014)Pennington, Socher, and
  Manning}]{pennington2014glove}
Jeffrey Pennington, Richard Socher, and Christopher~D. Manning. 2014.
\newblock Glove: Global vectors for word representation.
\newblock In \emph{EMNLP}.

\bibitem[{Pradhan et~al.(2012)Pradhan, Moschitti, Xue, Uryupina, and
  Zhang}]{pradhan2012conll}
Sameer Pradhan, Alessandro Moschitti, Nianwen Xue, Olga Uryupina, and Yuchen
  Zhang. 2012.
\newblock Conll-2012 shared task: Modeling multilingual unrestricted
  coreference in ontonotes.
\newblock In \emph{Joint Conference on EMNLP and CoNLL-Shared Task}.

\bibitem[{dos Santos and Zadrozny(2014)}]{santos2014:icml}
C\'{\i}cero~Nogueira dos Santos and Bianca Zadrozny. 2014.
\newblock Learning character-level representations for part-of-speech tagging.
\newblock In \emph{Proceedings of the 31st International Conference on Machine
  Learning (ICML), JMLR: W\&CP volume 32}.

\bibitem[{Soon et~al.(2001)Soon, Ng, and Lim}]{soon2001machine}
Wee~Meng Soon, Hwee~Tou Ng, and Daniel Chung~Yong Lim. 2001.
\newblock A machine learning approach to coreference resolution of noun
  phrases.
\newblock \emph{Computational linguistics}, 27(4):521--544.

\bibitem[{Turian et~al.(2010)Turian, Ratinov, and Bengio}]{turian2010word}
Joseph Turian, Lev Ratinov, and Yoshua Bengio. 2010.
\newblock Word representations: a simple and general method for semi-supervised
  learning.
\newblock In \emph{ACL}.

\bibitem[{Vilain et~al.(1995)Vilain, Burger, Aberdeen, Connolly, and
  Hirschman}]{vilain1995model}
Marc Vilain, John Burger, John Aberdeen, Dennis Connolly, and Lynette
  Hirschman. 1995.
\newblock A model-theoretic coreference scoring scheme.
\newblock In \emph{Proceedings of the 6th conference on Message understanding}.

\bibitem[{Wiseman et~al.(2016)Wiseman, Rush, and Shieber}]{wiseman2016learning}
Sam Wiseman, Alexander~M Rush, and Stuart~M Shieber. 2016.
\newblock Learning global features for coreference resolution.
\newblock In \emph{NAACL}.

\bibitem[{Wiseman et~al.(2015)Wiseman, Rush, Shieber, and
  Weston}]{wiseman2015learning}
Sam~Joshua Wiseman, Alexander~Matthew Rush, Stuart~Merrill Shieber, and Jason
  Weston. 2015.
\newblock Learning anaphoricity and antecedent ranking features for coreference
  resolution.
\newblock In \emph{ACL}.

\end{thebibliography}
\bibliographystyle{acl_natbib}
\end{document}